# Generative AI for automatic topic labelling

Diego Kozlowski[*], Carolina Pradier[**] and Pierre Benz[**]

[*]*diego.kozlowski@umontreal.ca*
0000-0002-5396-3471
School of Library and Information Science, University of Montreal, Canada

[**] *carolina.pradier@umontreal.ca; pierre.benz@umontreal.ca*
0009-0007-5058-6352; 0000-0003-1672-7311
School of Library and Information Science, University of Montreal, Canada

**Abstract**
Topic Modeling has become a prominent tool for the study of scientific fields, as they allow for a large scale interpretation of research trends. Nevertheless, the output of these models is structured as a list of keywords which requires a manual interpretation for the labelling. This paper proposes to assess the reliability of three LLMs, namely *flan*, *GPT-4o*, and *GPT-4 mini* for topic labelling. Drawing on previous research leveraging BERTopic, we generate topics from a dataset of all the scientific articles (n=34,797) authored by all biology professors in Switzerland (n=465) between 2008 and 2020, as recorded in the Web of Science database. We assess the output of the three models both quantitatively and qualitatively and find that, *first*, both GPT models are capable of accurately and precisely label topics from the models' output keywords. *Second*, 3-word labels are preferable to grasp the complexity of research topics.

## 1. Introduction

Topic models have become popular tools to identify the topics of scientific publications in an effort to map the ever-growing corpus of scientific knowledge (Blei & Lafferty, 2009; Suominen & Toivanen, 2016). However, the outputs of these unsupervised Natural Language Processing (NLP) algorithms can be challenging to interpret and evaluate, usually requiring manual validation by domain experts (Marchetti & Puranam, 2020; Rüdiger et al., 2022; Rijcken et al, 2023). Against this backdrop, delivering accurate and interpretable topic labelling is key. Indeed, state-of-the-art topic modeling techniques leveraging embeddings typically produce topic representations encompassing several terms, which are often hard to summarize into simple, informative topic labels. Hence, we propose to explore the potential of Large Language Models (LLM) to facilitate the interpretation of the output generated by topic models. Specifically, we compare the performance of *flan* (Chung et al., 2022) model by Google and Open AI's GPT 4 model (OpenAI et al., 2024) (in its full version, *GPT-4o*, and the *GPT-4 mini* version) in providing meaningful labels to complex models outputs.

Throughout this article we focus on the translation stage of topic assignment (Lancaster, 2003; Hjørland, 2017), i.e., the production of a topic label that is informative of the socially defined problem area within which each set of scientific publications particularly contributes to problem-solving (Hjørland, 1992). We use the output of a BERTopic model (Devlin et al., 2019) as our starting point and focus on producing labels for its topics in order to improve its interpretability. BERTopic is one of the most recent features of deep learning-based topic modeling (Grootendorst, 2022). BERTopic relies on BERT (Bidirectional Encoder Representations from Transformers) pre-trained embeddings to capture and represent the complexity of semantic relations in a corpus of text (Kang & Evans, 2020). Document embeddings are used to identify clusters of semantically similar documents. Each topic (cluster) therefore represents a group of documents that are measured as semantically similar. The topic representations in BERTopic are built using a modified version of the term frequency-inverse document frequency (TF-IDF) procedure in order to identify words that are highly representative of each document cluster—instead of highly representative of each document— (Grootendorst, 2022). This technique considers both how frequently a term appears within the cluster (in other words, the topic) with how rare it is across all clusters, selecting words that are simultaneously frequent in the cluster and infrequent in the rest of the corpus. The topic representation can be fine-tuned using models such as KeyBERTInspired, which selects among this set of keywords those that are most similar to the embedding of the most representative documents of each topic. However, the resulting topic representation (a set of words) usually remains hard to interpret and difficult to use in some tasks such as visualization of topics, for which hand labelling is a common practice.

This article is organized as follows: *first*, we present our topic model, as well as the prompts used to produce the topic labels using three generative language models. *Second*, we evaluate the models' performance relying on a set of quantitative and qualitative metrics. *Finally*, we present our main conclusions.

## 2. Methods

For our empirical evaluation, we use a dataset of all the scientific articles (n=34,797) authored by all biology professors in Switzerland (n=465) between 2008 and 2020, as recorded in the Web of Science database. In our previous work (Benz et al., 2024), we used this dataset to compare topics generated by LDA (Blei et al., 2003), and BERTopic (Grootendorst, 2022). We focus on this dataset because it generates a rather small topical space—104 topics—which, combined with our previous knowledge of the field, allows for a qualitative evaluation of the results. We use BERTopic with a minimum cluster size of 100 documents to create the base topic model and focus our analysis on the topic representation. Our strategy is to provide three different LLM with the ten keywords provided by BERTopic to represent each topic. We used the *flan* (Chung et al., 2022) model by Google, an open source generative language model; and Open AI's GPT 4 model (OpenAI et al., 2024) in its full version, *GPT-4o*, and the *GPT-4 mini* version. The *flan* model is run locally in Python using the *text2text* generation pipeline, while GPT models are used through OpenAI's API. For each of the three models, we tried two different prompts in order to produce first a single word label, and then a label with up to 3 words:

**Short name prompt:**

> *I have a corpus of Biology with 100 topics. I have a topic that is described by the following keywords: [KEYWORDS]*
>
> *Based on the information above, extract a short topic label of a single word that can accurately represent the topic, in the following format:*
>
> *topic: <topic label>*

**Long name prompt:**

> *I have a corpus of Biology with 100 topics. I have a topic that is described by the following keywords: [KEYWORDS]*
>
> *Based on the information above, extract a short topic label between one and three words that can accurately represent the topic, in the following format:*
>
> *topic: <topic label>*

Where [KEYWORDS] is the list of 10 keywords that represent the topic. Each model was run 20 times for each prompt. Our analysis is a combination of quantitative and qualitative metrics of performance. A sample of the resulting output is displayed in Table 1, and the metrics in use are defined in Table 2.

**Table 1. Exemplar topics**

| Type | Model | Topic 32 | Topic 77 |
|---|---|---|---|
| - | Keybert keywords | ['carbon', 'forests', 'co2', 'forest', 'ecosystem', 'vegetation', 'climate', 'biomass', 'ecosystems', 'photosynthesis'] | ['earthworms', 'earthworm', 'soil', 'soils', 'ecosystems', 'ecosystem', 'organisms', 'ecological', 'biodiversity', 'fungi'] |
| 1 word name | flan | ecosystem | earthworm |
| | GPT-4 mini | Carbonalogy | Soilfauna |
| | GPT-4 | carbon | soil |
| 1-3 words name | flan | biodiversity | earthworm |
| | GPT-4 mini | Carbon Ecosystems | Soil Ecosystems |
| | GPT-4 | Carbon Cycle in Forests | Soil Ecosystems |

**Table 2. Evaluation metrics**

| Metric | Range | Dimension | Notes |
|---|---|---|---|
| Number of distinct labels | (1,104) | Quantitative | For the 104 topics, how many different labels are created. |
| Stability | (1,20) | Quantitative | On the 20 iterations, how many different labels are created for the same topic |
| Similarity | (-1, 1) | Quantitative | The average cosine similarity on the embedding space of labels for a given topic between models. |
| Accuracy | (1,5) | Qualitative | How well does the label capture the content of the topic. |

Our quantitative evaluation is based on three metrics. First, the number of distinct labels created for the 104 topics is computed. Ideally, we aim to have a distinct label for each topic. We compute the average number of distinct labels across the 20 iterations and the 95% confidence interval. It is worth saying that, alternatively, repeated labels could be a sign of duplicated topics, but this would be a different use of generative models, where a prompt specific for the task of duplication detection or hierarchical grouping would be more useful. To measure the stability of the labels, we compute the number of different labels generated across the models for the same topic. Ideally, we want our models to be stable, and therefore a lower number is preferred. However, a model could create different labels that are nevertheless very similar; consequently, we complement this metric with the cosine similarity between embedding representations of all 20 labels generated for the same topic across iterations. We also compute the cosine similarity between models to quantify the differences between labels across models.

Our qualitative evaluation is based on the accuracy of the generated label to describe the topic. High scores in this metric imply that the label is *correct* in the sense that it reflects on the content of the topic, and *precise*, because its granularity adequately reflects the granularity of the topic. In this regard, the evaluation was conducted considering that some topics are more general than others. Table 3 displays information on qualitative coding and examples of varying quality labels for each topic in Table 1.

**Table 3. Qualitative coding**

| Score | Description | Interpretation | Topic 32 | Topic 77 |
|---|---|---|---|---|
| 1 | bad labelling | Inadequate to identify the topic | biology | biology |
| 2 | insufficient labelling | Globally refers to the topic but is too generic to provide a distinct categorization | ecosystem; carbon; carbonalogy* | soil |
| 3 | acceptable labelling | Indicative enough to identify the topic, but still lacking precision | biodiversity | earthworm; soilfauna |
| 4 | good labelling | Depicts the topic with significant precision | carbon ecosystems | - |

| Score | Description | Interpretation | Topic 32 | Topic 77 |
|---|---|---|---|---|
| 5 | perfect labelling | Very precisely describes the topics, outperforming labels with a score of 4 | carbon cycle in forests | Soil Ecosystems |

\* It is worth noting that, even though this concerns a very small minority of cases, it's possible for GPT to create neologisms like "carbonalogy".

For example, for the topics presented in Table 1, 'biology' as a label would get a score of 1, for it would not allow for any insight about the topic. 'Soil', 'carbon' and 'ecosystem' are typical representatives of a score of 2, for they are more precise, yet still too generic because they may refer to a variety of topics. 'Earthworm', 'soilfauna' and 'biodiversity' will get a score of 3, for they are indicative enough to ensure an understanding of the topic and usually specific enough to distinguish this topic from others. 'Carbon ecosystems' is an example for a score of 4, as its description is much more precise. Finally, 'carbon cycle in forests' is even more precise, and therefore gets a score of 5.

## 3. Results

As we are comparing an open source model, *flan*, and to paid models, considering the costs for each model is fundamental. Table 4 indicates the costs for our experiments on a rather small model (104 topics), run 20 times for each of the prompts, which is equivalent to a medium-large topic model (~4160 topics). Overall, as we only used the keywords of the topic and not the representative documents, the costs are rather small. Nevertheless, at the moment we run our experiments (August 2024), the GPT-4 model costs 30 times more than the mini version, which can be a relevant factor for larger implementations. Another note to take is that given that Open AI models are run through an API that require some seconds between each use, each iteration of each prompt and model took approximately ten minutes, while running the *flan* model was several times faster.

**Table 4. Costs on 20 iterations, 2 prompts, and 104 topics**

| Model | Cost | pricing |
|---|---|---|
| *flan* | Free | Free |
| *GPT-4* | $ 3.19 | $5.00 / 1M input tokens |
| *GPT-4 mini* | $ 0.11 | $0.150 / 1M input tokens |

Overall, we observe a big difference between the performance of the *flan* model, and the *GPT* models, but not a relevant difference between the *mini* and the regular *GPT-4* model. Figure 1A shows the number of unique labels by model and prompt type, for 104 topics. The closer to 104 a model is, the more distinct the labelling for each topic. For this metric, we observe that the *flan* model has a tendency to repeat labels across topics, which makes it unfit for the task. The length of the label also determines the generative models' capacity to create distinct labels;

when using a single word, repetitions are more frequent. With an almost perfect score across the 20 iterations, the *GPT-4 mini* is the best model in terms of distinctiveness.

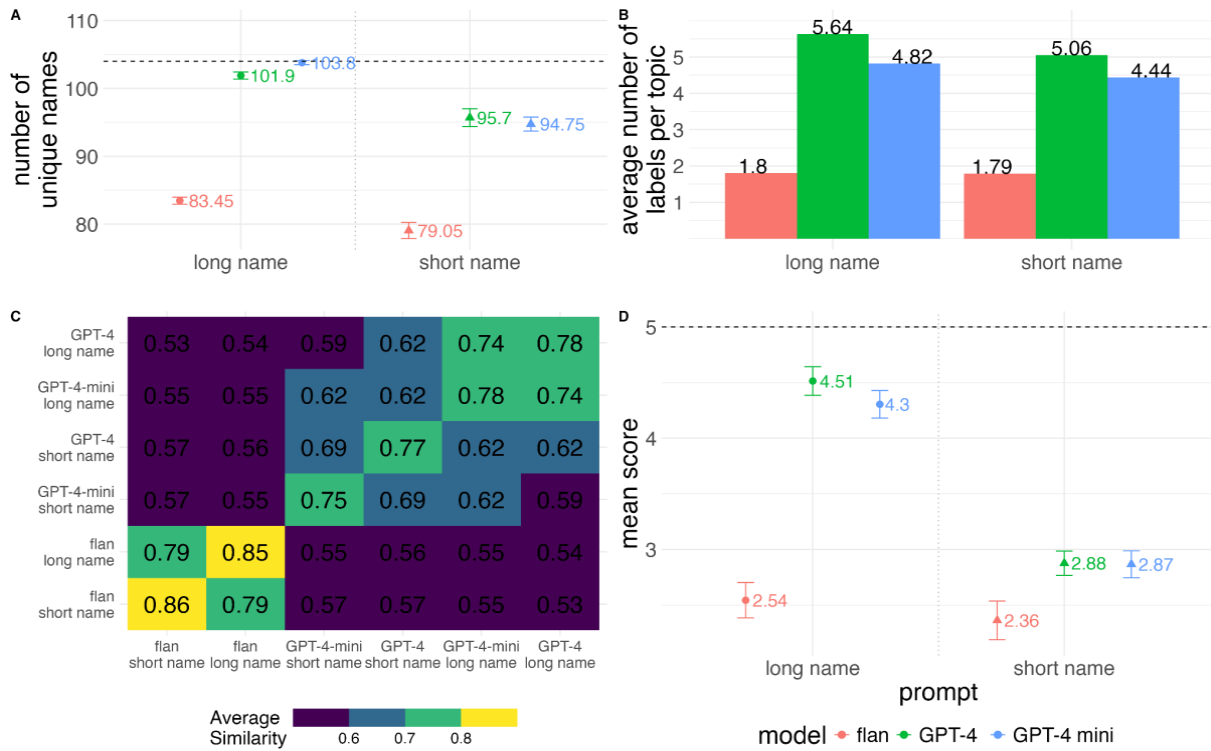

**Figure 1. Evaluation metrics.** Number of unique labels by model and prompt type (A), Average number of distinct labels per topic by model and prompt (B), cosine similarity between models (C), and score on the qualitative evaluation (D).

Figure 1B shows the average number of distinct labels per topic by model and prompt in 20 iterations. The closer to 1 the metric is, the more stable the model is. Given the tendency to repeat a smaller set of labels, the *flan* model is also more stable across different iterations. Short names are also more stable than longer names, but not to a large extent. Nevertheless, Figure 1B accounts for the exact repetition of the labels, but does not consider the similarity between alternative labelling across iterations. To tackle this, we computed the embedding representation of each label using the BERT *paraphrase-multilingual-MiniLM-L12-v2* model, and the cosine similarity among labels. For each model, we can compute the similarity with the other 19 iterations of the same topic and the average similarity. To compare different models, we compute the average cosine similarity for all the combinations of labels for each topic. Figure 1C shows the similarity matrix. Again, given the tendency to repeat itself, the *flan* model shows a high similarity across iterations. Also, given the tendency to create single word labels even when it was allowed to use up to three words, there is a high similarity between the two prompts of the *flan* model. The two versions of the *flan* show nevertheless a low similarity with all the *GPT* models. For this family of models and prompts, we see that there is a higher consistency among all of them, and the similarity increases when we use the same prompt. We also observe that the longer names are in general more similar than the shorter names, both for the same prompt and between prompts. This indicates that the longer names are able to capture a higher nuance that can be consistently reflected across iterations.

These results are consistent with the qualitative evaluation, as seen in Figure 1D. Our score, based on human evaluation of the labelling accuracy, is calibrated in such a way that a score of three is acceptable for use, while values below three are deemed unacceptable representations of the topics. The *flan* model scores equally poorly results for both prompts while the *GPT* models show a big difference between shorter and longer labels. Without a great variability, the short name prompt scores slightly below our acceptance threshold for both the *mini* and the standard version. On the other hand, the labels between one and three words are remarkably good, up to the point where human labelling by non-experts would probably underperform in such a task. Our qualitative evaluation also implies the exercise of thinking possible labels for the topic, and we come to realize that a single word is many times incapable of grasping the complexity of the topics. In this sense, the lower scores for the *short* prompt are related with the infeasibility of the task itself. Looking into more detail, we observe that the *GPT-4* model outperforms the *mini* model, but not significantly.

## 4. Discussion

Two major conclusions can be drawn from this exercise: first, the state of the art in generative modeling shows that models are already capable of accurately and precisely label research topics using only the keywords produced by the models' output. Second, to address this task, we need to use not less than three words in order to grasp the complexity of research topics.

Nevertheless, some nuance is necessary to contextualize these conclusions. Generative models' capacity to create synthetic labels is related with the overall quality of the topics themselves. Our previous work showed that BERTopic is able to create a good representation of the topic space for this same dataset (Benz et al., 2024). But this is not necessarily valid for all other corpus or models, and therefore the synthetic labels themselves could show a lower quality if the topics are not properly defined. In relation to this caveat, our analysis was carried out on a rather small corpus. A larger corpus (e.g. all Biology papers instead of those from Swiss professors) will tend to generate more topics, with a higher granularity. The capability of generative models to create distinctive labels for each topic on these larger spaces needs to be assessed. In this project, we followed the implementation of BERTopic for the AI labelling. In particular, this implies that when sending the request to the API, each topic is evaluated separately. The generative models do not have an overview of all the topics, and might result in imprecise or repetitive labels on larger topic spaces. A future line of work is to evaluate the LLM's performance on larger topic models, as well as the possibility of posterior disambiguation of equal labels for different topics. Finally, our open access benchmark model, *flan-t5-base*, is the one suggested by the BERTopic guide, but there are several alternatives that could outperform this model for the task. Large Language Models for *text2text* generation (the type of task performed on our experiments) are constantly being developed. We expect to see in the short future open access implementations that can achieve similar performance to the private versions.

**Open science practices**

In this project, we used the Web Of Science database to retrieve articles authored by biology professors in Switzerland. We decided to use this dataset because it is a well known case of study for the authors for the evaluation of Topic Models (Benz et al., 2024), which allows a qualitative evaluation of the results. The topics, with the different representations by the generative models and the scores are openly available at https://github.com/DiegoKoz/bertopic_gpt_labelling. The source code and all the results for the different iterations of the models are also available at the same repository.

**Acknowledgments**
Authors may include a brief acknowledgments section at the end of their paper to thank individuals and organisations that have supported their research.

**Author contributions**
Conceptualization: DK,PB, CP
Data curation:PB
Formal analysis: DK
Investigation: DK
Methodology: DK,PB
Software: DK
Supervision
Validation: PB
Visualization: DK
Writing – original draft: DK
Writing – review & editing, PB, CP

**Competing interests**
The authors have no competing interests.

**Funding information**
This project was funded by the Social Science and Humanities Research Council of Canada Pan-Canadian Knowledge Access Initiative Grant (Grant 1007-2023-0001), and the Fonds de recherche du Québec - Société et Culture through the Programme d'appui aux Chaires UNESCO (Grant 338828).
Pierre Benz acknowledges funding from the SNSF in the framework of its PostDoc.Mobility scheme (grant number: 210805).